\documentclass[conference]{IEEEtran}
\IEEEoverridecommandlockouts
\usepackage{cite}
\usepackage{subcaption}
\usepackage{amsmath,amssymb,amsfonts}
\usepackage{algorithmic}
\usepackage{graphicx}
\usepackage{textcomp}
\usepackage{xcolor}
\usepackage{booktabs}
\usepackage{multirow}
\usepackage{draftwatermark}
\def\BibTeX{{\rm B\kern-.05em{\sc i\kern-.025em b}\kern-.08em
    T\kern-.1667em\lower.7ex\hbox{E}\kern-.125emX}}
\begin{document}
\title{Post-Training Quantization for Energy Efficient Realization of Deep Neural Networks}

\author{\IEEEauthorblockN{Cecilia Latotzke, Batuhan Balim, and Tobias Gemmeke}
\IEEEauthorblockA{
Chair of Integrated Digital Systems and Circuit Design, RWTH Aachen University, 52062 Aachen Germany\\
Email: latotzke@ids.rwth-aachen.de
}
}

\SetWatermarkText{Accepted at 2022 21st IEEE International Conference on Machine Learning and Applications (ICMLA)}
\SetWatermarkColor[gray]{0.75}
\SetWatermarkFontSize{0.4cm}
\SetWatermarkAngle{0}
\SetWatermarkHorCenter{11cm}
\SetWatermarkVerCenter{27cm}
\maketitle

\begin{abstract}
\textcolor{black}{The biggest challenge for the deployment of Deep Neural Networks (DNNs) close to the generated data on edge devices is their size, i.e., memory footprint and computational complexity.
Both are significantly reduced with quantization. 
With the resulting lower word-length, the energy efficiency of DNNs increases proportionally.
However, lower word-length typically causes accuracy degradation.
To counteract this effect, the quantized DNN is retrained.
Unfortunately, training costs up to 5000$\times$ more energy than the inference of the quantized DNN.
To address this issue, we propose a post-training quantization flow without the need for retraining.
For this, we investigated different quantization options.
Furthermore, our analysis systematically assesses the impact of reduced word-lengths of weights and activations revealing a clear trend for the choice of word-length.
Both aspects have not been systematically investigated so far. 
Our results are independent of the depth of the DNNs and apply to uniform quantization, allowing fast quantization of a given pre-trained DNN.
We excel state-of-the-art for 6\thinspace bit by 2.2\% Top-1 accuracy for ImageNet.
Without retraining, our quantization to 8\thinspace bit surpasses floating-point accuracy.}
\end{abstract}


\section{Introduction}

Deep Neural Networks (DNNs) excel in classification tasks and achieve better classification accuracies than other algorithms \cite{ImageNetChallenge}.
Driving factors for the algorithmic performance of DNNs are their number of layers (depth) and width of channels, which result in an increasing number of weights and activations as well as in increasing computational complexity typically expressed as a number of Multiply-Accumulate operations (MAC).
Both, the number of weights and activations, proportionally drive the cost in terms of memory and computations.
The number of MAC and the related need for communication to storage largely impact the energy consumption for classifying data with DNNs \cite{Horowitz2014}.
\textcolor{black}{In Table \ref{tab:energy_memory} the most common memory storage technologies and there corresponding energy\thinspace/\thinspace bit are listed.
Most commonly, DDR3 is used as memory technology for weights and input image storage in low-power neural network accelerators \cite{Latotzke2021}.
}

The memory footprint is proportional to the word-length of weights and activations.
Typically, weights and activations are represented in 32\thinspace bit floating-point during training.
Quantization is one of the most effective techniques when it comes to energy reduction \cite{Latotzke2021} as it concurrently reduces the signal paths to memory and within the processing units.
Thereby, quantization refers to the step of defining a reduced word-length in bits as well as specifying the unit of least precision. Equivalently, a scaling factor can be specified assuming a normalized data range.
\begin{table}[htbp]
\addtolength{\tabcolsep}{-0.9pt}
\caption{\textcolor{black}{Energy for memory accesses}}
\label{tab:energy_memory}
\centering
\begin{tabular}{|c|c|c|c|c|}
\hline
                & DDR3        & LPDDR3        & DDR4     & SRAM       \\ \hline
Reference       & \cite{Malladi2012} & \cite{Schaffner2015} & \cite{Akhlaghi2018} & \cite{Clerc2012} \\ 
\hline
Memory near to logic   & no          & no            & no          & yes        \\ \hline
Energy          & 70\thinspace pJ\thinspace/\thinspace bit    & 21\thinspace pJ\thinspace/\thinspace bit      & 15\thinspace pJ\thinspace/\thinspace bit   & 55\thinspace fJ\thinspace/\thinspace bit   \\ \hline
\end{tabular}
\end{table}

In the case of embedded devices, private data or low latency aware data has to be processed on low-power devices, while maintaining classification accuracy \cite{Mittal2020}.
However, accuracy may suffer from word-length reduction 
\cite{Yang2019}.
To minimize the impact of quantization, it can be accounted for in training leading to the so-called Quantization Aware Training (QAT).
\textcolor{black}{QAT means that after quantization during the inference, the network is trained with floating-point precision during backpropagation.
This is often called retraining or finetuning.
Although many methods for retraining-based quantization achieve good accuracy results \cite{Esser2019, Bhalgat2020, Bai2021}, they cost immense additional energy and time. 
Most commonly, training is done by means of a GPU like RTX 2080 TI, which is using between 140\thinspace W to 350\thinspace W \cite{igor}. We assume for now 150\thinspace W.
For this, one training epoch of ResNet-50 on ImageNet in 16\thinspace bit floating-point with an average of 466 frames\thinspace/\thinspace s needs 386\thinspace kJ\thinspace/\thinspace epoch \cite{lambdablog}.
Retraining takes between 100 to 200\thinspace epochs \cite{Esser2019, Bai2021}.
This results in 38.6\thinspace MJ to 77.2\thinspace MJ for QAT on ImageNet compared to 16\thinspace kJ for inference of the complete ImageNet validation set.
Finally QAT takes more time with 71.5\thinspace hours to 143\thinspace hours compared to inference of 107\thinspace s.}

Alternatively, Post-Training Quantization (PTQ) applies quantization on a parameter set that was trained in a machine representation.
With the lack of additional retraining, this approach risks significant degradation in accuracy as compared to the floating-point training parameter set.
\textcolor{black}{At the same time, producing a new quantized parameter set is done with comparatively very low effort.
Hence, an optimized PTQ should preserve the accuracy of the training parameter set as much as possible while being almost 5000$\times$ faster and more energy efficient than the QAT approach.
}

The process of quantization enforces a scaling of the precision and number range. 
Many works explicitly consider a related scaling factor.
However, there has not been any systematic investigation for the best way of determining this factor.

This paper proposes a novel PTQ flow including the following contributions:
\begin{itemize}
\item Results and discussion of a systematic evaluation of quantization options considering the highest achievable accuracy of a uniformly quantized DNN with reduced word-length.
Our analysis identifies the best combinations of different quantization options. 
\item Our investigated quantization options cover the computation methods of scaling factors and the baseline distributions for scaling factor computation. 
Proposed computation methods of scaling factors (AbsMax and AbsP) provide better results than state-of-the-art metrics used to determine weight scaling factors and activation scaling factors.
Furthermore, we identified the benefit of channelwise over layerwise computation of weight scaling factors regardless of DNN depth and word-length.
\item The results of our systematic study indicate that preserving a higher word-length of weights as opposed to activations provides better accuracy which contradicts existing hypotheses \cite{Sze2019}. 
\item Furthermore, we show that the Mean Square Error (MSE) of the quantization 
is an unreliable predictor of achieved accuracy when used to benchmark different quantization techniques.
\item Finally, our PTQ flow provides an accuracy - energy and accuracy - memory footprint trade-off and achieves for selected quantization options higher accuracy than the respective floating-point baseline.
\end{itemize}

The paper is organized as follows.
In Section \thinspace\ref{sec:relatedWorks} a brief overview of the related work and background is provided.
Section\thinspace\ref{sec:Methods} introduces the applied methods.
Section\thinspace\ref{sec:Experiments} details the experimental settings. 
The results are presented in Section\thinspace\ref{sec:results}. The paper concludes with Section\thinspace\ref{sec:conclusion}.

\section{Related Works}
\label{sec:relatedWorks}
\textcolor{black}{
There exist a variety of compression techniques for DNNs to increase efficiency during inference \cite{Deng2020, Gholami2021}.
Most prominent are pruning \cite{Dong2017} and quantization of weights or activations \cite{Han2016}.
Pruning usually needs retraining of the network to maintain accuracy, due to its invasive nature.
This retraining needs a significant amount of energy, as earlier discussed.
In this work, we focus on a compression method without retraining: post-training quantization, PTQ.}
In general, most works applying PTQ, adopt a uniform word-length for all layers, round the quantized values to the nearest, use a default scaling factor, and apply a uniform quantization pattern.
To counteract the resulting accuracy drop, some works apply an adaptive quantization pattern, which is a data-driven approach.
Thereby, the set of quantized values follows the distribution of the original data more closely by adapting the quantization step accordingly.
In one approach, it is modulated as a function of the weight distribution \cite{Nogami2019}.
Another work relies on the position of the most significant non-zero digit in a binary number \cite{Gupta2020}.
Here, the layerwise distribution is used to identify an adaptive log-two-based quantization pattern while rounding numbers up.

To improve PTQ accuracy the commonly used round to nearest can be replaced by an adaptive rounding technique \cite{Nagel2020}.
It is based on the layerwise statistics of weights and aims to minimize the MSE between the quantized weights and the floating-point weights.
The scaling factor is predefined on a layerwise basis by minimizing the MSE of the floating-point weights and the round to the nearest quantized weights. 
\textcolor{black}{
Another approach to minimize the MSE is to compute the expected MSE per layer before it propagates through the network \cite{Nagel2019}.
Unfortunately, this work is limited to quantization from 32\thinspace bit floating-point to 8\thinspace bit fixed-point.}

Furthermore MSE is used to compute sensitivity metrics which aim to model the impact of quantization per layer \cite{Lee2021}.
In general, it is assumed that the sensitivity per layer towards disturbances like noise is layer-dependent.
Hence, an ideal mixed-precision word-length should exist for each DNN. 
This could be identified by predicting the expected signal-to-quantization-noise-ratio per layer \cite{Lin2016} or by using an individual word-length per layer, i.e., mixed-precision, to reduce the expected induced error by quantization for each layer \cite{Zhou2018}.
\textcolor{black}{An even more granular approach compared to layerwise quantization is channelwise quantization \cite{Krishnamoorthi2018}. 
Here, the impact of symmetric and asymmetric quantization patterns per layer and per channel on accuracy were investigated.
}

So far, there has not been an investigation of scaling factor computation methods for PTQ.
Nor has there been an in-depth analysis of the word-length impact of weights or activations for PTQ other than a brief investigation on CIFAR10 and MNIST \cite{Mitschke2019}.
Since these small datasets provide only limited challenges to DNNs, an in-depth analysis of the word-length in this work adopts the ImageNet dataset.
\textcolor{black}{Furthermore, we are the first to systematically investigate scaling factor computation methods. 
For this, we apply QAT based methods to PTQ and add two, so-far not introduced methods in literature, to compute the scaling factor. 
Finally, we are the first to show that MSE is not a useful metric for DNN quantization.}
\vspace{-0.2cm}

\section{Methodology}
\label{sec:Methods}

In the following, three key aspects of quantization are introduced: 
\begin{itemize}
\item The fundamental mapping from the set of precise $\Gamma_\text{FP}$ (\textbf{F}loating \textbf{P}oint) to the set of reduced precision $\Gamma_\text{Q}$ (\textbf{Q}uantized). Thereof, the latter is typically a subset of the former $\Gamma_\text{Q} \subset \Gamma_\text{FP}$. The mapping is commonly realized as a scaling operation followed by saturation, cf. Section\thinspace\ref{sec:quantization}.
\item MSE as commonly used metric to locally evaluate the quality of quantization, cf. Section\thinspace\ref{sec:MSE}.
\item Different approaches to define the aforementioned scaling. 
This includes the scaling factor computation and the statistical baseline for weight scaling factor computations, cf. Section\thinspace\ref{sec:QuantOpt}. 
\end{itemize}

\subsection{Quantization}
\label{sec:quantization}

\begin{figure}[htbp]
\centering
\includegraphics[scale=1]{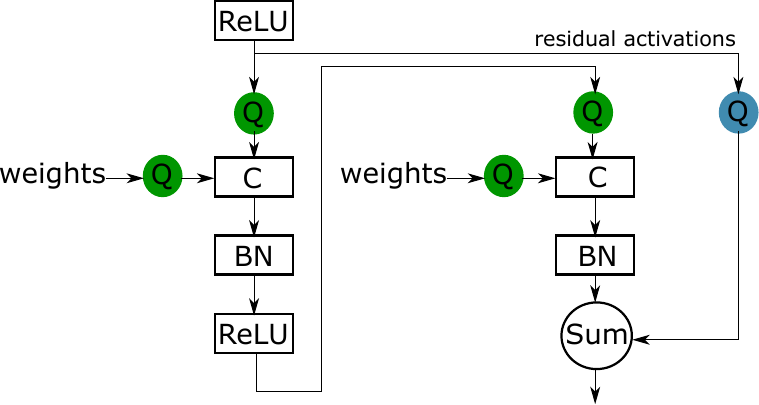}
\caption{Scheme of a single residual block with convolution C, Batch Norm BN, Rectified Linear Unit ReLU, and quantization node Q. The Quantization of residual activations is optional, indicated by a blue node.}
\label{fig:quantNodes}
\end{figure}

In the following, the quantization is realized by applying a scaling factor to map the numeric values from the precise set to the quantized set of numbers. 
The scaled values are then rounded and saturated to the specified number representation of the target set.
For quantization of a floating-point value to a lower precision fix-point value, a target word-length and a quantization scaling factor are needed. 
We quantize floating-point DNNs from Torchvision without retraining using quantization nodes Q (cf. Fig.\thinspace\ref{fig:quantNodes}).

For quantization, we focus on uniform quantization because of its common use in literature.
Hence, the results are widely applicable. 
In the following, FP, Q, and INT abbreviate floating-point, quantized, and integer, respectively.
\begin{equation}
\label{eq:quant}
x^\text{Q} = x^\text{INT} \times s
\end{equation}
\begin{equation}
\label{eq:int}
x^\text{INT} = \lfloor(\text{clamp}(\frac{x^{\text{FP}}}{s}, x^\text{INT}_\text{min}, x^\text{INT}_\text{max})\rceil
\end{equation}

Applying uniform quantization to floating-point values limits the possible range and precision of the so-called quantized fix point values $x^\text{Q}$, which are represented as product of an integer $x^\text{INT}$ and the scaling factor $s$ as in Eq.\thinspace(\ref{eq:quant}). 
To adjust the range between the smallest, $x^\text{INT}_\text{min}$, and the largest, $x^\text{INT}_\text{max}$, and the unit of least precision, only two factors are needed: the chosen word-length $b$ per weight or activation and the corresponding scaling factor $s$. As the step size is assumed to be uniform, it is identical to the unit of least precision. 
Activations are unsigned based on the use of ReLU as non-linear activation function, hence, for activations $a^\text{INT}_\text{min}$ = 0 and $a^\text{INT}_\text{max}$ = $2^{b}-1$.
Weights are quantized as signed numbers, so, for weights $w^\text{INT}_\text{min}$ = $-2^{b-1}$ and $w^\text{INT}_\text{max}$ = $2^{b-1}-1$.
The round to nearest function is indicated with $\lfloor\cdot\rceil$, clamp$(\cdot)$ limits all values exceeding its lower and upper bounds, $x^\text{INT}_\text{min}$ and $x^\text{INT}_\text{max}$.
We apply Eq.\thinspace(\ref{eq:quant}) with Eq.\thinspace(\ref{eq:int}) for activations and weights replacing $x$ with the respective variable.

\subsection{Mean Square Error of the Quantization}
\label{sec:MSE}

Many studies utilize the MSE to optimize the hyper-parameters of the quantization, e.g., \cite{Uhlich2019,Bai2021, Bhalgat2020}.
Based on $N$ full precision values ${x}^\text{FP}_i$ and the corresponding quantized values ${x}^\text{Q}_i$, the MSE is computed according to Eq.\thinspace(\ref{eq:MSE}).
\begin{equation}
    MSE = \frac{1}{N}\sum_{i=0}^{N-1}({x}^\text{FP}_{i} - {x}^\text{Q}_{i})^2
\label{eq:MSE}
\end{equation}

\subsection{Quantization Options}
\label{sec:QuantOpt}

Quantization options contain the baseline distribution (i.e., layerwise or channelwise distribution) to compute scaling factors and the scaling factor computation methods for weights (WSM) and activations (ASM).
\subsubsection{Baseline Distribution}
weight scaling factors could be either computed based on the layerwise or channelwise weight distribution.
As it is common, activation scaling factors are always computed based on layerwise activation distribution.

\subsubsection{Computation of scaling factors}

The scaling factor computation LSQ in \cite{Esser2019} uses the mean $\langle\cdot\rangle$ of the absolute values $|\cdot|$ of the weights or activations, Eq.\thinspace(\ref{eq:LSQ}). 
\begin{equation}
\label{eq:LSQ}
s_\text{LSQ} = \displaystyle\frac{2\langle|\textbf{x}|\rangle}{\sqrt{{x^\text{INT}_\text{max}}}}
\end{equation}

Weights are expected to be Gaussian distributed for LSQ+ \cite{Bhalgat2020}, with the mean being represented by {$\mu$} and the standard deviation being represented by $\sigma$, Eq.\thinspace(\ref{eq:LSQ+}). 
\begin{equation}
\label{eq:LSQ+}
s_\text{LSQ+} = \displaystyle\frac{\max(|{\mu} -3\sigma|,|{\mu} +3\sigma|)}{|{{w^\text{INT}_\text{min}}}|}
\end{equation}

The study of extreme value computations in \cite{Bai2021} suggests that the best computation uses the minimum and maximum activation $a$ per channel and computes the mean over all channels $C$, cf. Eq.\thinspace(\ref{eq:Bquant}).
The computation of the mean over all DNN dimension but the channel-width, is indicated by $i_\text{B},i_\text{H},$ and $i_\text{W}$. 
Here, B, H, and W represent batchsize, feature map height, and feature map width, respectively. 
\begin{equation}
\label{eq:Bquant}
s_\text{BatchQuant} = \displaystyle\frac{\frac{1}{C}\sum\limits_{i_\text{C}}^{C}({\displaystyle\max_{i_\text{B},i_\text{H}, i_\text{W}}(\textbf{a}_{i_\text{C}})}-{\displaystyle\min_{i_\text{B},i_\text{H}, i_\text{W}}(\textbf{a}_{i_\text{C}})})}{{a^\text{INT}_\text{max}}-{a^\text{INT}_\text{min}}}
\end{equation}

\textcolor{black}{Additionally to the state-of-the-art scaling factor computation methods, we present two methods, which are so far not introduced in literature.
These methods compute} the scaling factor based on the statistical distribution of weights or activations with regards to either their absolute maximum `AbsMax' cf. Eq.\thinspace(\ref{eq:AbsMax}), or absolute percentile `AbsP' cf. Eq.\thinspace(\ref{eq:AbsP}).
For the percentile function, per$_k()$ the hyperparameter $k$ determines which upper percentile of the maximum values is chosen, i.e., $\max()$ equals per$_k()$ with $k = 100$.
Possible $k$-values are between [0, 100] with $k = 99.99$ leading to best accuracy in our study.
\begin{equation}
\label{eq:AbsMax}
s_\text{AbsMax} = \displaystyle\frac{\max (|\textbf{x}|)}{{{x^\text{INT}_\text{max}}}}
\end{equation}
\begin{equation}
\label{eq:AbsP}
s_\text{AbsP} = \displaystyle\frac{\text{per}_{k} (|\textbf{x}|)}{{x^\text{INT}_\text{max}}}
\end{equation}

\subsubsection{Accuracy analyses}

To compare the set of quantization options across varying word-lengths, we introduce the figure of merit $acc_\text{diff}$ that captures the deviation in the accuracy of a specific quantization concerning the mean of all experiments using the word-length, cf. Eq.\thinspace(\ref{eq:acc}).
Here, $N$ represents the total number of criteria and wl represents the word-length.
\begin{equation}
\label{eq:acc}
acc_\text{diff} = \underbrace{acc^\text{wl}_{\text{c}}}_{\text{accuracy for criterion $c$ with wl}} - \underbrace{\frac{1}{N}\sum_{c}^{criteria}{acc}^\text{wl}_{{c}}}_{\text{mean accuracy for wl}}
\end{equation}

\section{Experiments}
\label{sec:Experiments}

In our experiments we apply the ResNet models\cite{ResNet}: ResNet-18, -34, -50, and -152 on ImageNet \cite{ImageNet}.
They use a similar structure but increase in depth.
Hence, they allow a systematic comparison of the orthogonal dimensions: word-length combinations, scaling factor computation methods for weights and activations, the baseline distribution, and the sensitivity analysis of quantization of residual activations.
As it is practiced in state-of-the-art, accuracy refers to validation accuracy in the following and all images of the validation set are used for inference.
Our analysis focuses on accuracy during inference without retraining.

\subsubsection{word-length Combinations}

The word-length for all weights is swept in 5 steps from 4\thinspace bit to 8\thinspace bit and the word-length for all activations from 4\thinspace bit to 8\thinspace bit.
The large number of scaling factor computation methods for weights and activations is limited to the statistical methods, AbsMax and AbsP.
Furthermore, channelwise or layerwise weight distribution, and floating-point residual activations (fpRes) or quantized residual activations (qRes) are applied.
This results in 16 different settings per word-length combination and 25 different word-length combinations per DNN.

\subsubsection{Sensitivity Analysis}

We analyze the specific impact of quantization in the residual blocks by making it optional as shown in Fig.\thinspace\ref{fig:quantNodes}.
So, we either preserve the floating-point residual activations (fpRes) or quantize them (qRes).
Based on the assumption that for word-length below 6\thinspace bit accuracy is largely degraded, we limit in this analysis the evaluated range of word-lengths to the range from 6\thinspace bit to 8\thinspace bit.
At the same time, we apply identical word-lengths to weights and activations.

\subsubsection{Quantization Options}

For the analysis of the quantization options, i.e., the baseline distribution and the computation of scaling factors, we also use the minimum word-length of 6\thinspace bit.
Hence, we apply the same word-length for weights and activations in the range 6\thinspace bit to 8\thinspace bit.
The scaling factors $s$ are computed based on a sample of 1000 random images of the ImageNet validation set.

In the course of our experiments, we explore all permutations of WSM and ASM except for LSQ+ and BatchQuant.
LSQ+ was only applied to weight scaling factors in \cite{Bhalgat2020} and BatchQuant was only applied to activation scaling factors in \cite{Bai2021}.
LSQ as well as AbsP and AbsMax are applied to weight scaling factors as well as activation scaling factors.
To conclude, the application of a specific method to scaling factors of weights or activations is in accordance with the related publication suggesting the method.

\section{Results}
\label{sec:results}

Initially, the quality of MSE is evaluated concerning its capability to predict overall accuracy.
Then we present an analysis of the impact of word-lengths for weights and activations, the sensitivity analysis, and the benchmarking of quantization options.
Finally, we conclude by presenting the best combinations and comparing the results to the state-of-the-art.

\subsection{MSE as an indicator for model accuracy}

\begin{figure}[htbp]
\centering
\includegraphics[scale=0.42]{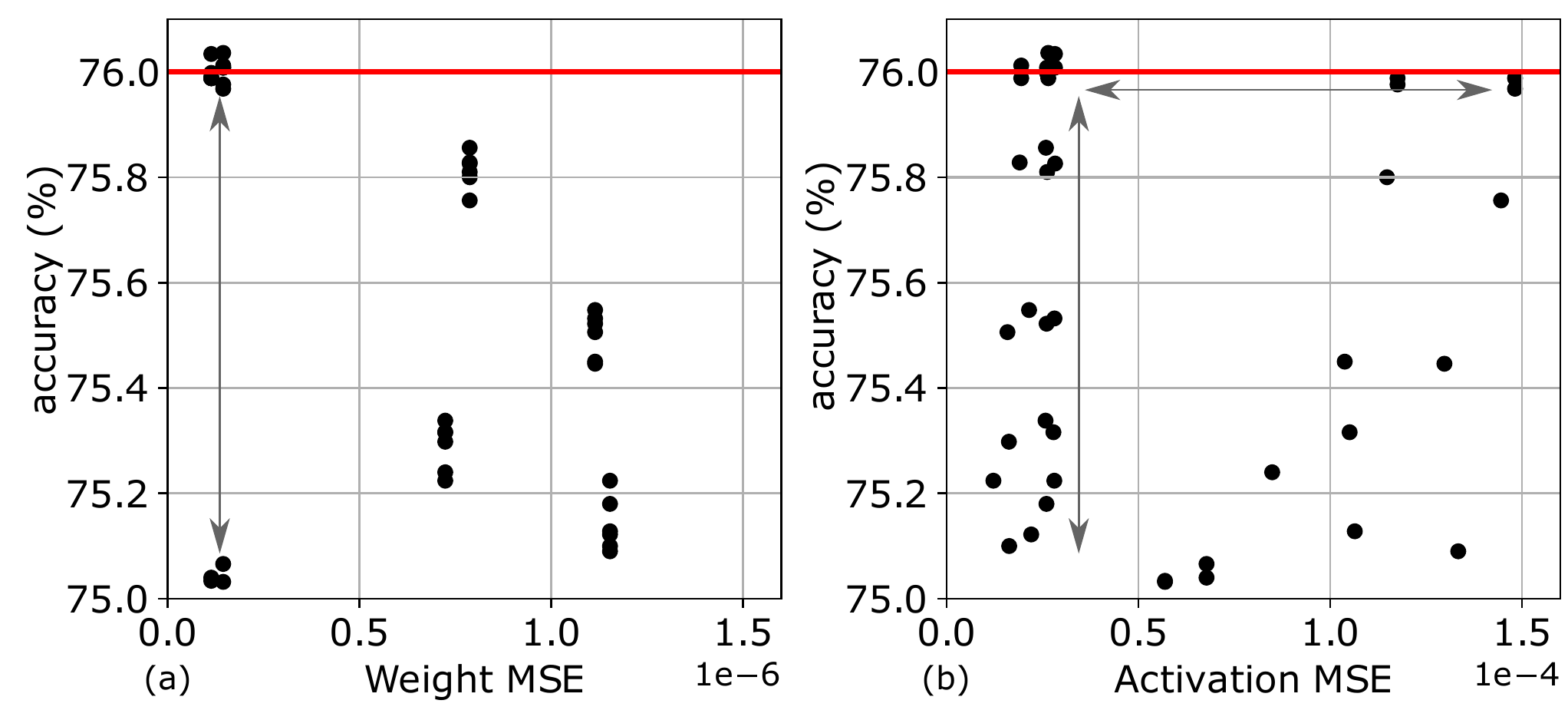}
\caption{Accuracy versus MSE for ResNet-50: (a) weight MSE and (b) activation MSE.
Weight and activation word-length is 8\thinspace bit. The floating-point baseline accuracy is shown in red.}
\label{fig:MSEvsAcc}
\vspace{-0.25cm}
\end{figure}

The accuracy of a quantized network is shown in Fig.\thinspace\ref{fig:MSEvsAcc} as a function of MSE as incurred due to quantization.
Here and in the following, ResNet-50 is used as a representative example of the ResNet-type network.
Results of other ResNet networks follow the same trends.

As was pointed out by other studies, quantization leads in some cases to higher accuracy than the corresponding floating-point baseline.
Considering the MSE of quantized weights in the left plot, quantization options that lead to comparable low MSE values feature a variation of 1\,\% in accuracy.
Using otherwise unchanged quantization options, activation MSE on the right is in general two orders of magnitude higher than the weight MSE.
A similar observation as on the left is made, as similar small MSE correspond to accuracy levels being spread by around 1\,\% accuracy.
Furthermore, even 10$\times$ larger MSE values correspond to quantization that produces accuracies within the top group.

To conclude, MSE appears not well correlated to achievable accuracy when applied across different quantization options.
Hence, using MSE is unsuited as a metric to select an appropriate quantization scheme.

\subsection{Analysis of word-length impact on accuracy}

\begin{figure}[htbp]
\centering
\includegraphics[scale = 0.65]{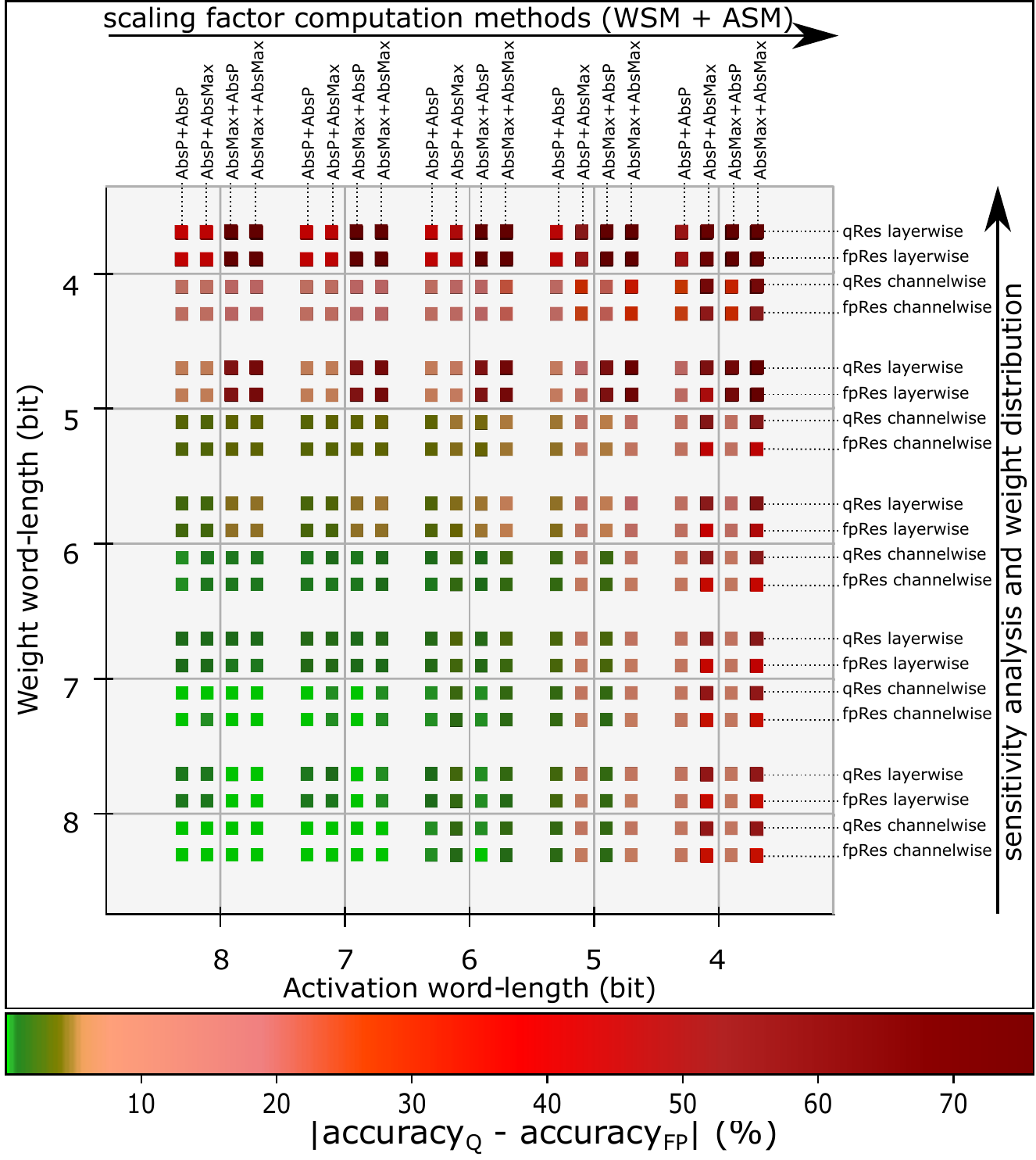}
\caption{Impact of weight and activation word-length for ResNet-50. 
Each word-length combination covers 16 data points, due to the combination of scaling factor computation methods, sensitivity analyses, and weight distribution. The color indicates the accuracy difference between the floating-point and quantized ResNet-50.}
\label{fig:AccvsWL_4to8bit_birdsview} 
\vspace{-0.3cm}
\end{figure}

In Fig.\thinspace\ref{fig:AccvsWL_4to8bit_birdsview} the accuracy loss reaches up to 70\,\% for 4\thinspace bit weights or activations.
Even for 5\thinspace bit, the accuracy drop is significant. 
It turns out that 5\thinspace bit activations show overall better results than 5\thinspace bit weights.
This challenges the common wisdom that activations need more precision than weights to maintain accuracy.
A reason might be the asymmetric distribution of quantized values considering only unsigned numbers for activations but signed numbers in the case of weights.
So, the single-sided distribution of activations has half the quantization step size providing better resolution for the same word-length as compared to the case of weights.
Hence, the reduction of weight word-length impacts accuracy more than the reduction of activation word-length.

\label{sec:wordlength}

\begin{figure*}[htbp]
     \centering
     \begin{subfigure}[b]{0.24\textwidth}
         \centering
\includegraphics[width=1\linewidth]{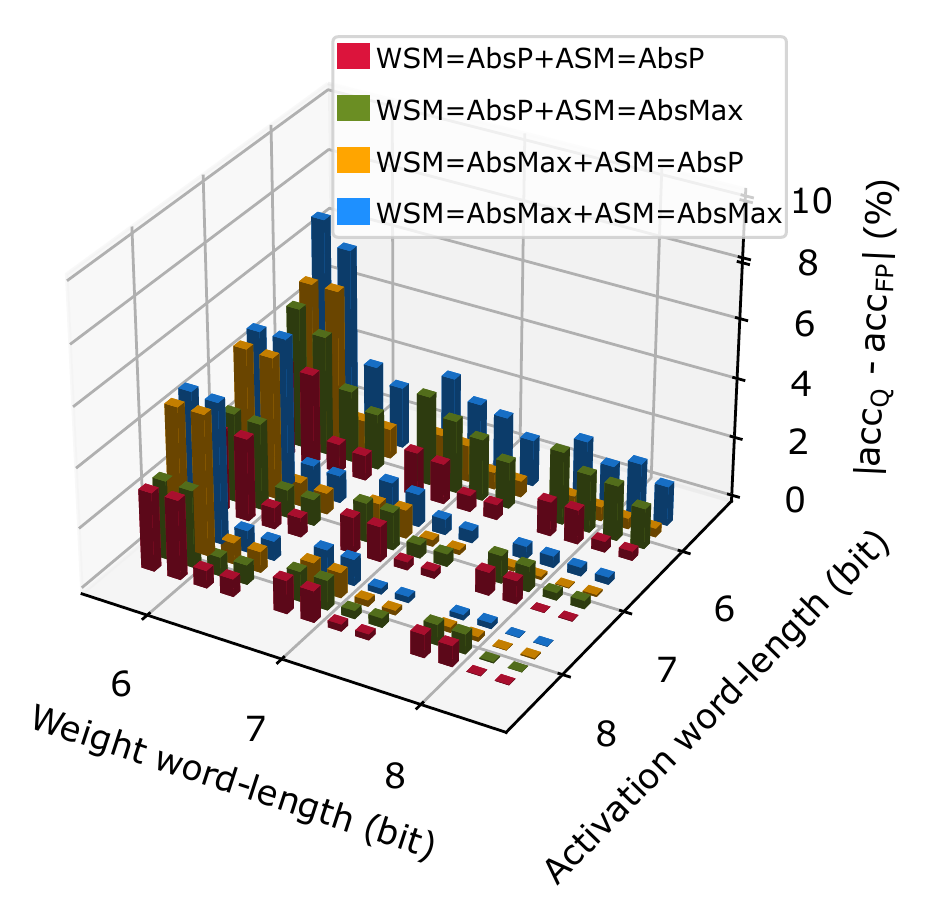}
\caption{}
\label{}
     \end{subfigure}
     \begin{subfigure}[b]{0.24\textwidth}
\includegraphics[width=1\linewidth]{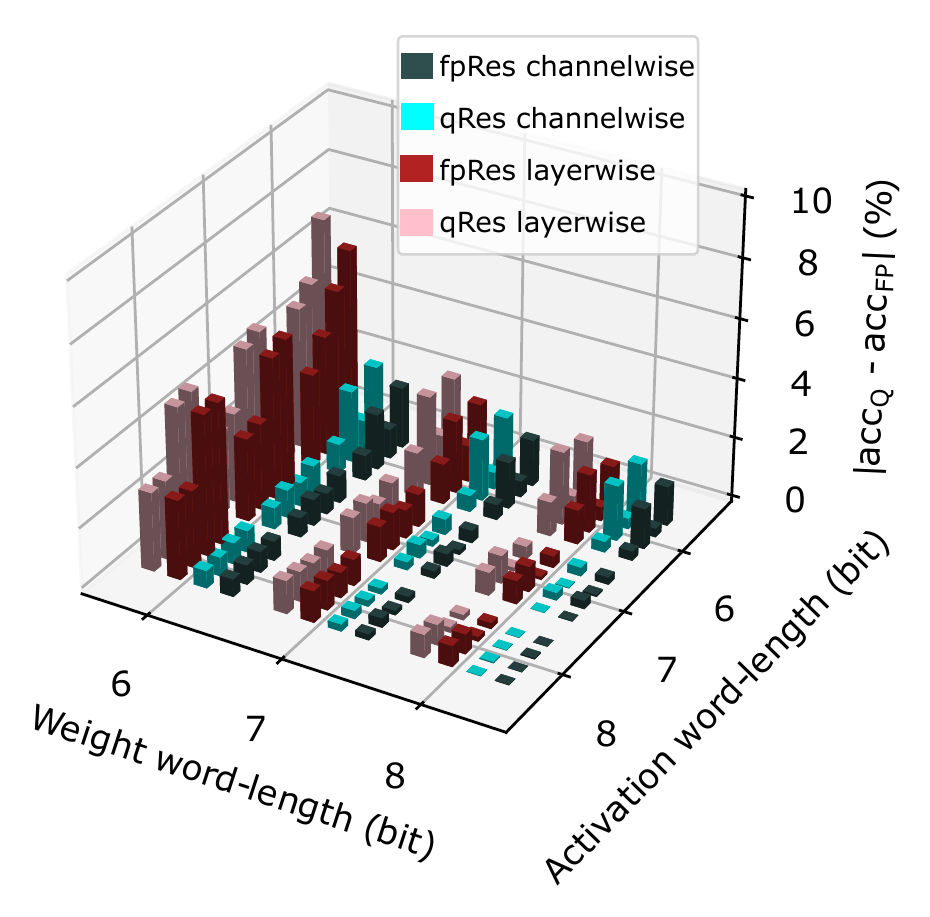}
\caption{}
\label{fig:AccvsWL_Opt}
     \end{subfigure}
     \begin{subfigure}[b]{0.24\textwidth}
\includegraphics[width=1\linewidth]{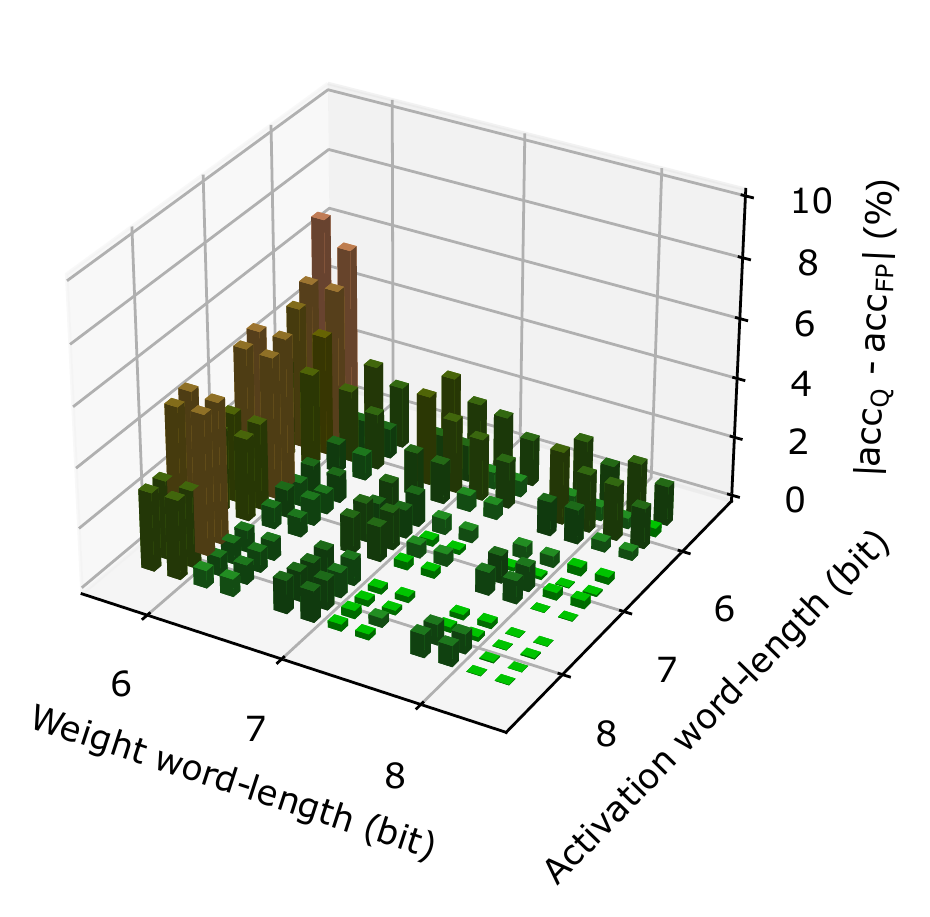}
\caption{}
\label{fig:AccvsWL_6to8bit}
     \end{subfigure}
     \begin{subfigure}[b]{0.24\textwidth}
\includegraphics[width=1\linewidth]{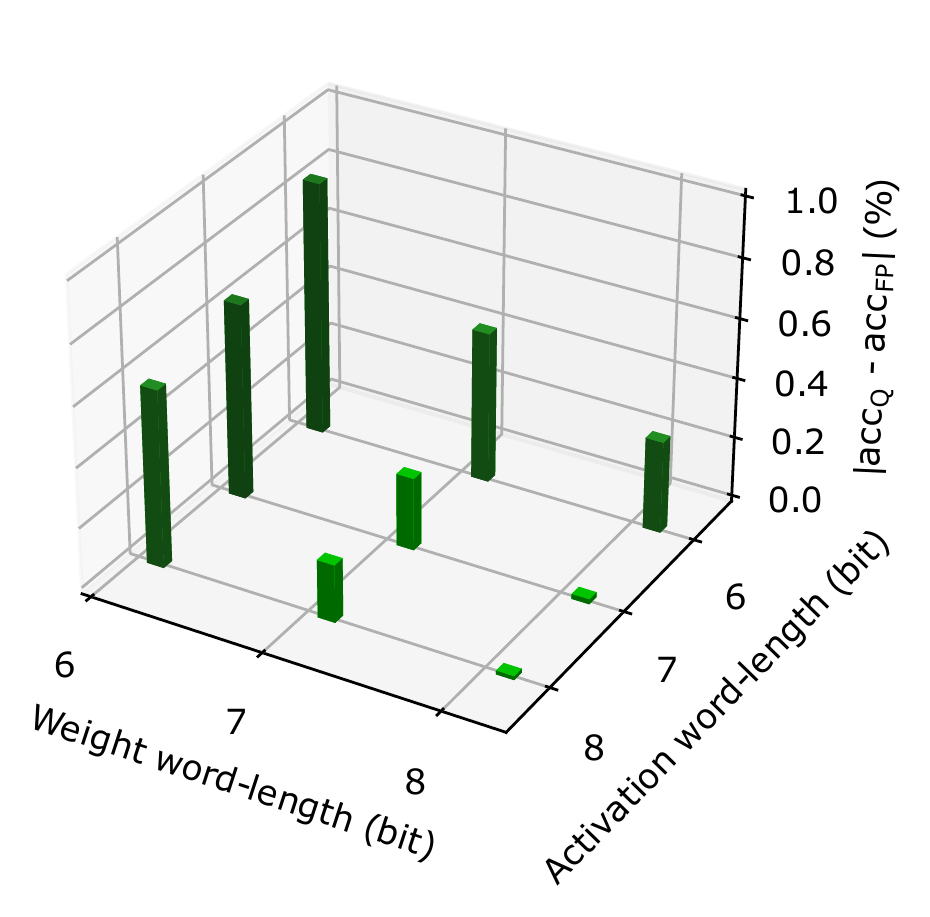}
\caption{}
\label{fig:AccvsWL_6to8bit_best}
     \end{subfigure}
        \caption{A zoom into the data presented in Fig.\thinspace\ref{fig:AccvsWL_4to8bit_birdsview} for ResNet-50 with 6\thinspace bit to 8\thinspace bit. Scaling factor computation methods (a) and sensitivity analysis and baseline distributions (b) are indicated by color, accuracy is indicated by bar height (c).
        In (d) the best performing combination, with floating-point residual activations (fpRes), channelwise as baseline distribution, and scaling factor computations methods WSM equal to AbsP and ASM equal to AbsP, is shown. Here, accuracy is indicated by the same color scheme as in Fig.\thinspace\ref{fig:AccvsWL_4to8bit_birdsview}.}
        \label{fig:WL-3D}
\end{figure*} 

In the following, we limit the evaluation on the word-lengths between 6\thinspace bit to 8\thinspace bit to assure a drop in accuracy of no more than 10\,\%, cf. Fig.\thinspace\ref{fig:WL-3D}.
Here, the absolute difference in accuracy of a quantized DNN acc$_\text{Q}$ and its respective floating-point accuracy acc$_\text{FP}$ is visualized by bar height.
Even for 6\thinspace bit, the impact of lower weight word-length is visible, especially for layerwise computation of the weight scaling factors.
Computing weight scaling factors per channel decreases the impact of precision loss by lower word-length.
For activation scaling factors, AbsMax yields lower accuracies than AbsP.
Finally, the best combination is shown in Fig.\thinspace\ref{fig:AccvsWL_6to8bit_best} that combines fpRes with channelwise computed weight scaling factors and WSM as well as ASM employing AbsP. 
This combination with 8\thinspace bit weights and 7\thinspace bit activations achieves an accuracy of 76.06\,\%, i.e., 0.06\,\% better than the respective floating-point baseline.

\subsection{Analysis of quantization options}

The impact on the accuracy of different quantization options is visualized in histograms.
The combination of word-length choices, sensitivity analysis, and quantization options creates a convoluted design space.
In the following, one of those aspects is considered at a time.

\label{sec:scaleMethodresults}
\begin{figure*}[htbp]
    \centering
    \begin{subfigure}[b]{0.32\textwidth}
        \includegraphics[width=1\linewidth]{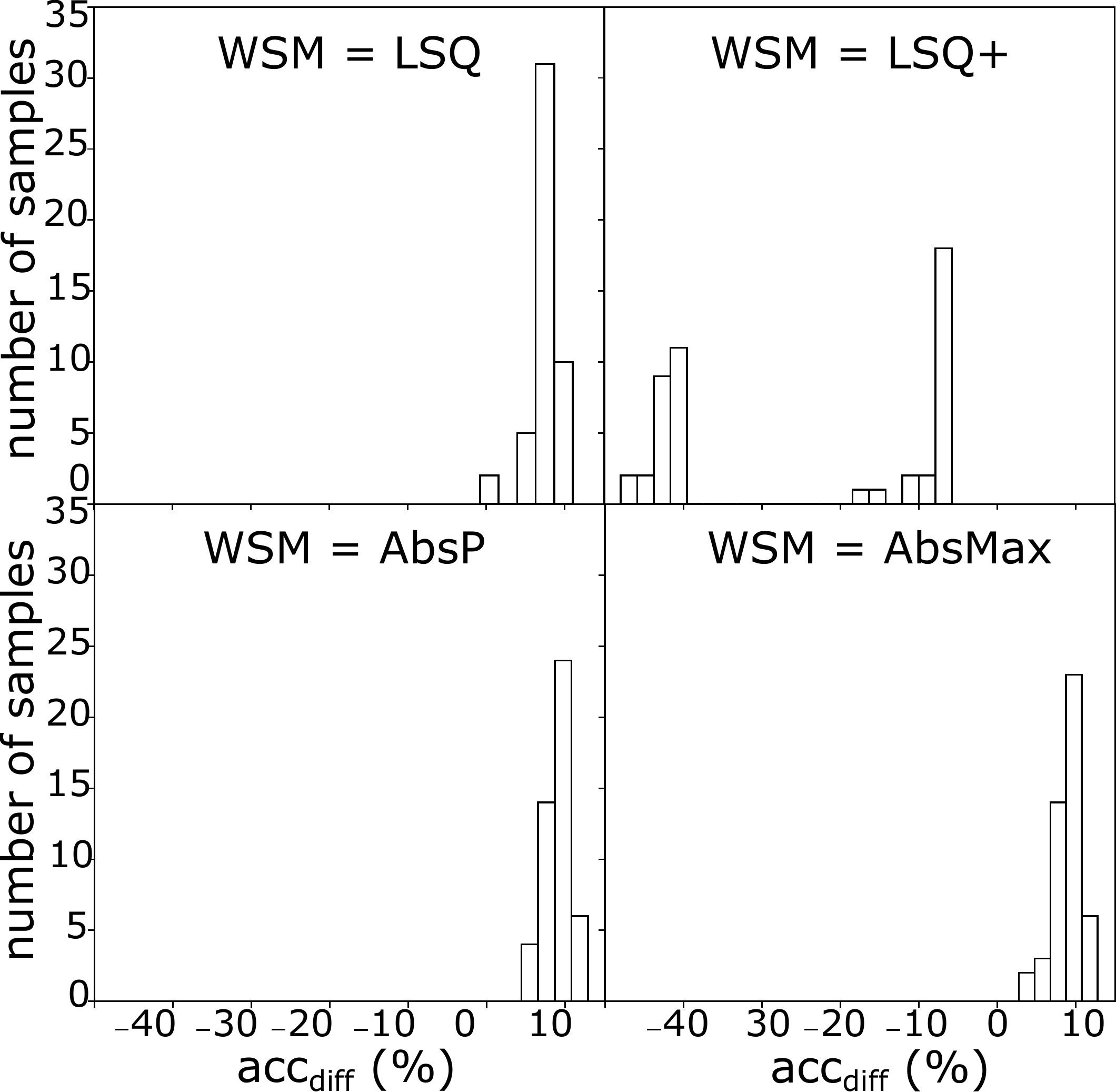}
        \caption{c = weight scaling factor computation methods (WSM)}
        \label{fig:WSM}
    \end{subfigure}
    \begin{subfigure}[b]{0.32\textwidth}
        \includegraphics[width=1\linewidth]{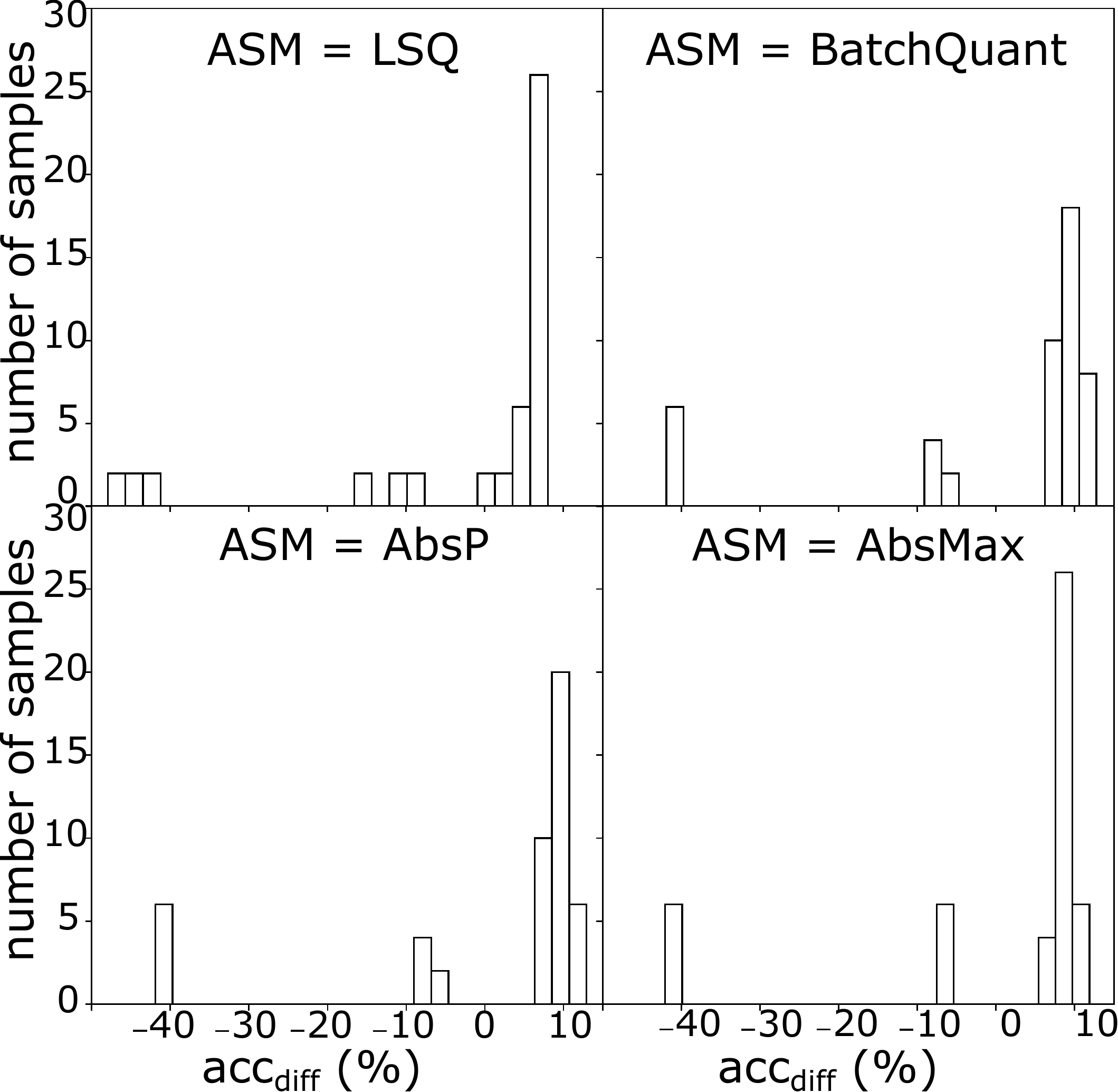}
        \caption{c = activation scaling factor computation methods  (ASM)}
        \label{fig:ASM}
    \end{subfigure}
    \begin{subfigure}[b]{0.32\textwidth}
        \includegraphics[width=1\linewidth]{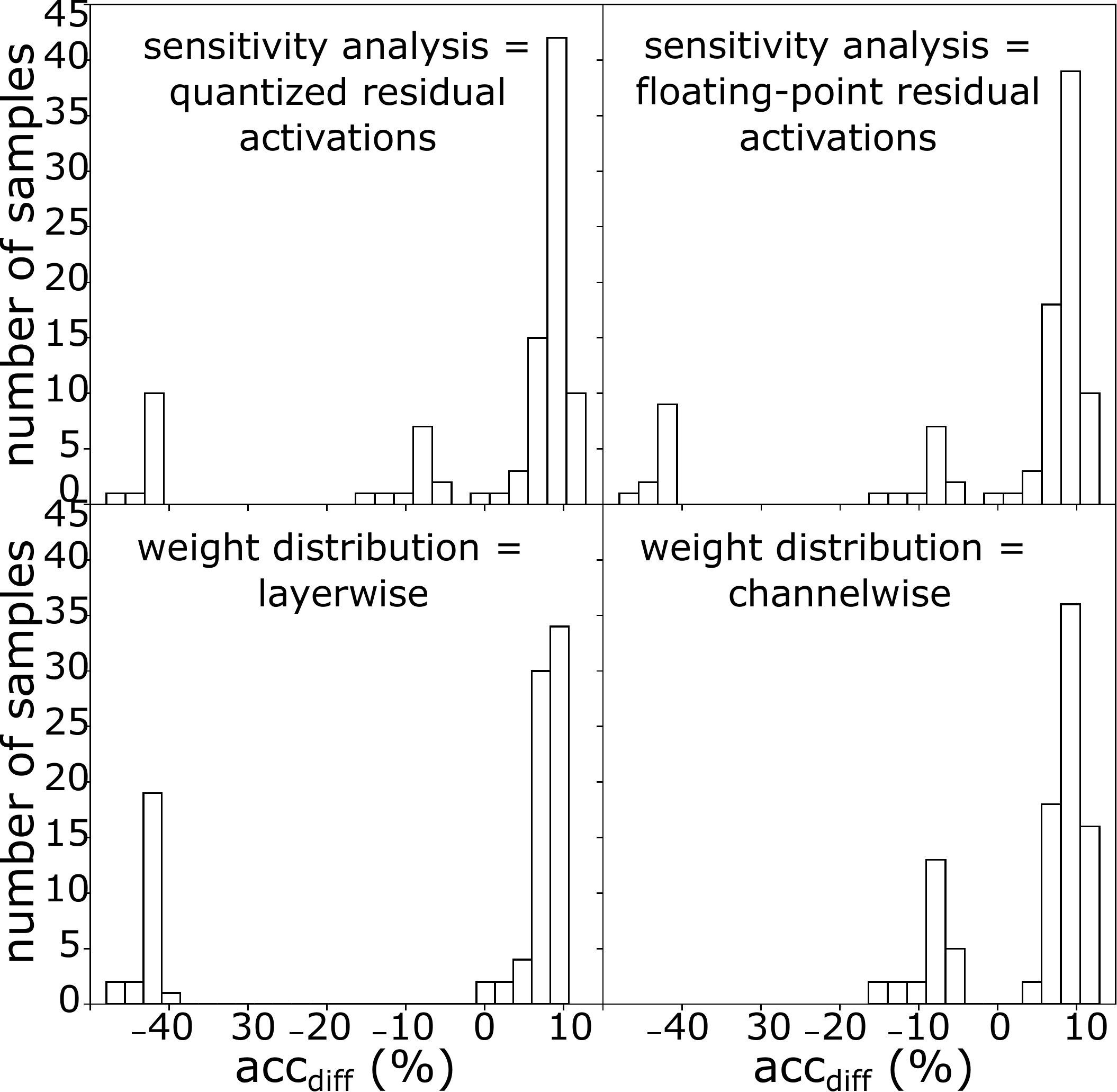}
        \caption{c = sensitivity analysis or c = weight distribution}
        \label{fig:Opt}
    \end{subfigure}
    \caption{Histogram for ResNet-50, all weights and activations are represented by similar word-length in the range of 6\thinspace bit to 8\thinspace bit. 
    Each combination of quantization options for a certain word-length is a sample.
    The bins contain samples with the according criteria stated in the subplot. 
    The difference of the accuracy `acc$_\text{diff}$' for an analyzed criteria compared to its mean word-length accuracy is computed by Eq.\thinspace(\ref{eq:acc}).}
    \label{fig:all_Hist}
    \vspace{-0.4cm}
\end{figure*}

\begin{table}[htbp]
\addtolength{\tabcolsep}{-4pt}
\centering
\caption{{Best quantization options for ImageNet with weight and activation word-length (wl$_\text{w}$ and wl$_\text{a}$)}}
\label{tab:bestMetOpt}
\begin{tabular}{@{}ccccccc@{}}
\multicolumn{1}{c}{\textbf{ResNet-}} & \multicolumn{1}{c}{{\textbf{wl$_\text{w}$}}/{\textbf{wl$_\text{a}$}}} & \multicolumn{1}{c}{\textbf{residual}}&\multicolumn{1}{c}{\textbf{baseline}}&  \textbf{WSM} & \textbf{ASM} & \multicolumn{1}{c}{\textbf{accuracy}} \\
                                 &             \textbf{(bit)}              & \textbf{activations} & \textbf{distribution} &   &   &                   \textbf{Top-1 (\%)}                 \\ \midrule
\multirow{5}{*}{\textbf{18}}     & FP/FP                         & -                  & -                   & -                             & -                             & \textbf{69.69}                              \\
                                 & 8/8                           & qRes               & channel             & AbsP                          & AbsMax                        & 69.62                              \\
                                 & 7/7                           & fpRes              & channel             & AbsMax                        & AbsP                          & 69.50                              \\
                                 & 6/6                           & qRes               & channel             & AbsP                          & AbsP                          & 68.48                              \\
                                 & 5/5                           & fpRes              & channel             & AbsMax                        & BatchQuant                        & 63.21                              \\ \midrule
\multirow{5}{*}{\textbf{34}}     & FP/FP                         & \textbf{-}         & \textbf{-}          & \textbf{-}                    & \textbf{-}                    & \textbf{73.27}                              \\
                                 & 8/8                           & qRes               & channel             & AbsP                          & AbsP                          & 73.24                              \\
                                 & 7/7                           & fpRes              & channel             & AbsMax                        & AbsP                          & 73.02                              \\
                                 & 6/6                           & fpRes              & channel             & AbsP                          & AbsP                          & 72.52                              \\
                                 & 5/5                           & fpRes              & channel             & AbsP                          & BatchQuant                        & 69.61                              \\ \midrule
\multirow{5}{*}{\textbf{50}}     & FP/FP                         & \textbf{-}         & \textbf{-}          & \textbf{-}                    & \textbf{-}                    & 76.00                              \\
                                 & \textbf{8/8}                           & \textbf{fpRes}              & \textbf{channel}             & \textbf{AbsMax}                        & \textbf{AbsP}                          & \textbf{76.04}                     \\
                                 & 7/7                           & fpRes              & channel             & AbsMax                        & AbsP                          & 75.84                              \\
                                 & 6/6                           & fpRes              & channel             & AbsP                          & AbsP                          & 75.15                              \\
                                 & 5/5                           & fpRes              & channel             & AbsP                          & BatchQuant                        & 71.12                              \\ \midrule
\multirow{5}{*}{\textbf{152}}    & FP/FP                         & \textbf{-}         & \textbf{-}          & -                             & -                             & 78.26                              \\
                                 & \textbf{8/8}                           & \textbf{fpRes}              & \textbf{channel}             & \textbf{AbsP}                          & \textbf{AbsMax}                        & \textbf{78.28}                     \\
                                 & 7/7                           & fpRes              & channel             & AbsMax                        & AbsP                          & 78.08                              \\
                                 & 6/6                           & fpRes              & channel             & AbsP                          & AbsP                          & 77.39                              \\
                                 & 5/5                           & fpRes              & channel             & AbsP                          & AbsP                          & 74.02                             
\end{tabular}
\end{table}

All three histograms in Fig.\thinspace\ref{fig:all_Hist} are based on the same experimental data set.
Each depicts the data according to the criterion $c$ mentioned in each subplot.
The impact of word-length is neutralized by showing only the deviation for each data point against the mean accuracy for the used word-length, cf. Eq.\thinspace(\ref{eq:acc}).

For WSM in Fig.\thinspace\ref{fig:WSM}, AbsMax and AbsP achieve higher accuracies than LSQ and LSQ+.  
The state-of-the-art computation of scaling factor methods LSQ, LSQ+, and BatchQuant benefit from the retraining of quantized DNNs.
Since in our experiment no training is applied, the purely statistical methods, AbsP, and AbsMax, to compute of scaling factor outperform the state-of-the-art methods.
For ASM, Fig.\thinspace\ref{fig:ASM} shows that LSQ performs worse than BatchQuant, AbsP, and AbsMax. 
AbsP and AbsMax offer less computational effort than BatchQuant while providing similar performance.
As a result, AbsP and AbsMax are the preferable ways to compute scaling factors for weights and activations.
Both are used for WSM and ASM in the analysis for word-length impact, cf. Section\thinspace\ref{sec:wordlength}.

Fig.\thinspace\ref{fig:Opt} visualizes the results of the sensitivity analyses and the impact of weight distribution on the accuracy.
It is worth noting that if the channelwise weight distribution is applied, layerwise weight distribution cannot be applied.
The same goes for the sensitivity analysis of residual activations against quantization, i.e., fpRes and qRes.
Comparing fpRes and qRes, no significant difference in accuracy is visible.
The statistical baseline options do provide a significant difference between channelwise and layerwise.
Channelwise is by far superior to layerwise.
This is caused by its finer adjustment to the distribution of weights.

\subsection{Settings of Best Performing DNNs}

Our detailed analysis of the impact of quantization options results in an overview of differently sized ResNets with different word-length combinations.
We record in Table\thinspace\ref{tab:bestMetOpt} the best accuracy for equal word-length for weights wl$_\text{w}$ and activations wl$_\text{a}$ and the respective quantization options.
The respective 32\thinspace bit floating-point baseline is indicated as FP.

For deeper DNNs, floating-point residual activations (fpRes) are superior compared to quantized residual activations (qRes) throughout different word-length combinations.
A reason could be that they preserve information from preceding layers, which would be otherwise lost.
The deeper a network gets the more relevant this effect becomes.

Computing weight scaling factors per channel is superior to per layer, since the adjustment of scaling factors is more adaptive to the distribution of weights.

In shallow to medium deep DNNs with low word-lengths, 5\thinspace bit or less, BatchQuant achieves best results as ASM.
It offers a trade-off between increased computational effort while maintaining accuracy for a lower memory footprint.  
However, for the most part, AbsMax and AbsP are the best computation methods of scaling factors for ASM and WSM.
They achieve the highest accuracies and require less computational effort than BatchQuant.
As a result, AbsP and AbsMax are the preferable computation of scaling factor methods.

Finally, with our suggested methods, floating-point accuracy could be surpassed, as indicated in bold.
Hence, a quantized DNN without additional training achieves higher accuracy than its full precision counterpart.
So, quantization noise can improve accuracy.

\subsection{State-of-the-art comparison}

\begin{figure}[htbp]
    \centering
    \begin{subfigure}[b]{0.45\textwidth}
        \includegraphics[width=1\linewidth]{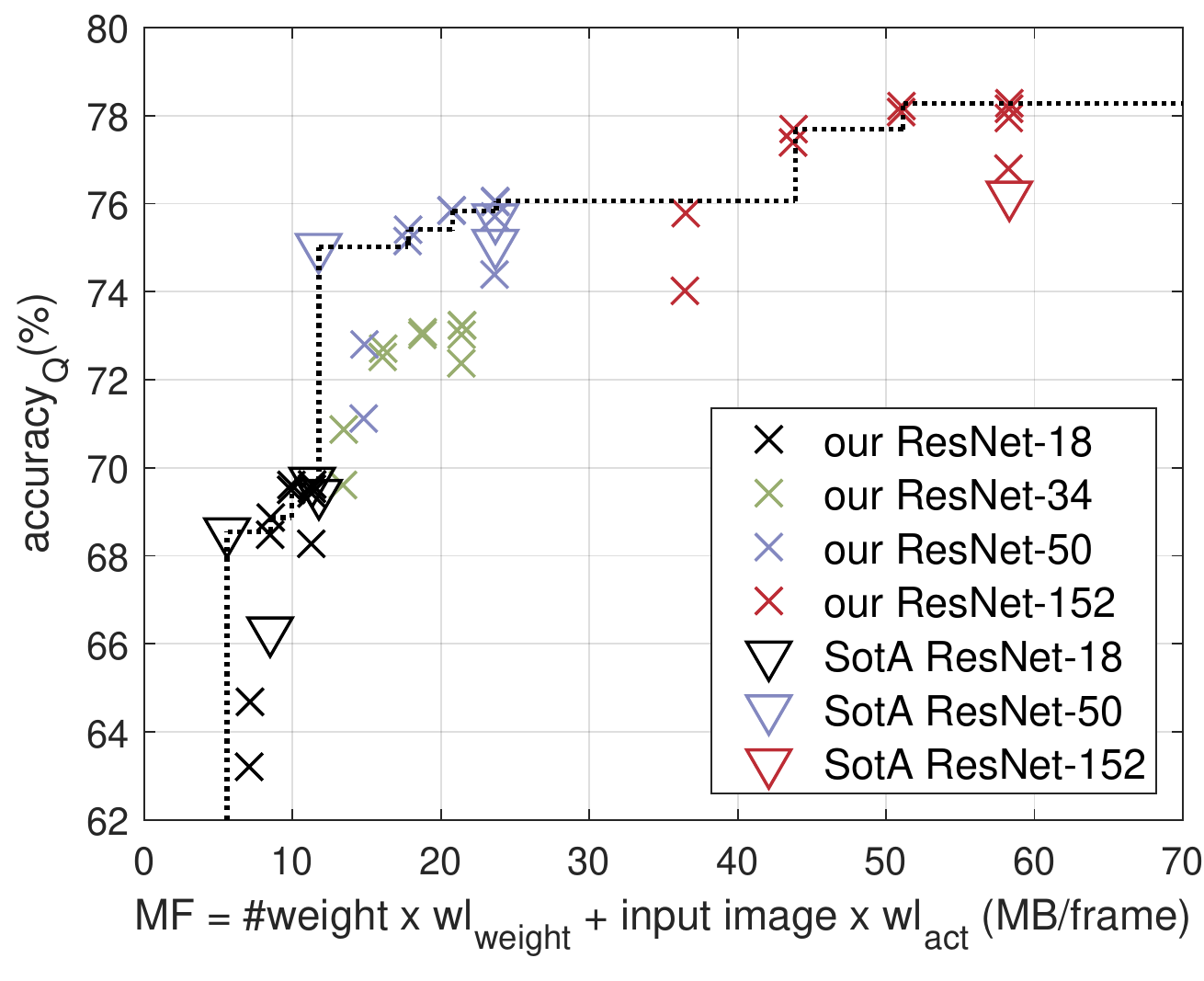}
        \caption{}
        \label{fig:MF}
    \end{subfigure}
    \begin{subfigure}[b]{0.45\textwidth}
        \includegraphics[width=1\linewidth]{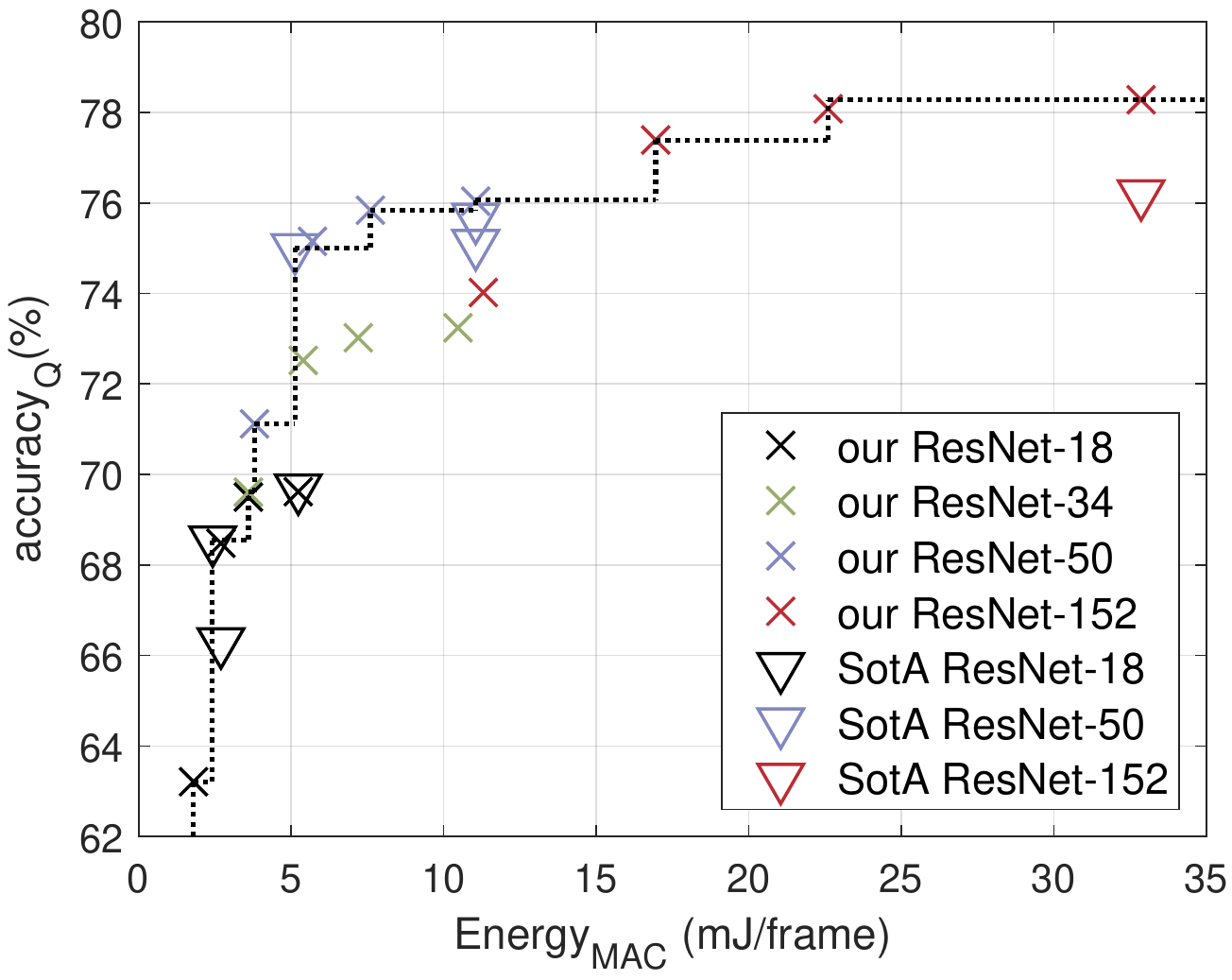}
        \caption{}
        \label{fig:MAC_energy}
    \end{subfigure}
    \caption{\textcolor{black}{The state-of-the-art (SotA) comparison in PTQ for accuracy$_\text{Q}$ versus (a) memory footprint (MF) and (b) MAC energy. 
    The Pareto front (black dotted line) highlights the optimal solutions.}}
    \label{fig:MF-CC}
    \vspace{-0.5cm}
\end{figure}

\textcolor{black}{
Fig.\thinspace\ref{fig:MF-CC} shows the optimal trade-off between cost and accuracy, i.e., the Pareto front.
Cost refers here either to the memory footprint MF, cf. Fig.\thinspace\ref{fig:MF} or the energy needed for the majority of computations in a DNN, which are MAC, cf. Fig.\thinspace\ref{fig:MAC_energy}. 
For MAC energy, we assume a deployment on an edge device with MAC computation units in 22\thinspace nm with energy values reported in \cite{Stadtmann2020}.
The floating-point baseline ResNets and their respective costs and accuracies are depicted in Table\thinspace\ref{tab:FP_resnet}.
The comparison between the floating-point baseline and the quantized DNNs results in more than 4$\times$ memory footprint and 27$\times$ energy reduction while maintaining floating-point accuracy.
Finally, the Pareto front in Fig.\thinspace\ref{fig:MF-CC} contains almost no medium deep DNNs but rather deeper DNNs with lower word-length, cf. compare ResNet-34 to ResNet-50.}

\begin{table}[htbp]
\addtolength{\tabcolsep}{-3.3pt}
\centering
\caption{\textcolor{black}{Memory footprint and energy for ResNet 32\thinspace bit floating-point baseline}}
\label{tab:FP_resnet}
\begin{tabular}{|c|c|c|c|c|}
\hline
                              & \textbf{ResNet-18} & \textbf{ResNet-34} & \textbf{ResNet-50} & \textbf{ResNet-152} \\ \hline
\textbf{Accuracy (\%)}        & 69.69              & 73.27              & 76.00              & 78.26               \\ \hline
\textbf{MF (MB/frame)}        & 45.29              & 85.73              & 94.68              & 233.23              \\ \hline
\textbf{Energy$_\text{MAC}$ (mJ/frame)} & 139.68             & 279.36             & 294.88             & 876.88              \\ \hline
\end{tabular}
\end{table}

The state-of-the-art comparison for works considering PTQ is shown in Table\thinspace\ref{tab:SotA}.
Here, the weight word-length wl$_\text{w}$ and activation word-length wl$_\text{a}$ are given in bit and the respective 32\thinspace bit floating-point baseline is indicated as FP.
All layers apply the same word-length.

\textcolor{black}{Our work outperforms state-of-the-art in preserving floating-point accuracy for quantized networks, cf. \cite{Krishnamoorthi2018} for ResNet-50 and ResNet-152 in Table\thinspace\ref{tab:SotA}.
Even though the floating-point baseline of \cite{Nagel2019} is even 0.01\% higher than ours, we excel with our quantized ResNet-18 for 6\thinspace bit \cite{Nagel2019} by 2.18\%.
For ResNet-50, we achieve 0.36\,\% higher accuracy than \cite{Lee2021} by similar word-length.
Our method surpass state-of-the-art, \cite{Krishnamoorthi2018} by 1.58\% for ResNet-152.
Finally, our work achieves between 0.02\,\% to 0.04\,\% higher accuracy for the quantized ResNet-152 and ResNet-50 compared to the respective floating-point baseline.
}
\begin{table}[htbp]
\addtolength{\tabcolsep}{-2.5pt}
\centering
\caption{State-of-the-art comparison for PTQ methods for ImageNet Top-1 accuracy 
}
\label{tab:SotA}
\begin{tabular}{c|cll|cl|cl}
\textbf{} & \multicolumn{3}{c|}{ResNet-18}                                                                  & \multicolumn{2}{c|}{ResNet-50}                                      & \multicolumn{2}{c}{ResNet-152}                           \\\hline
wl$_\text{w}$         & \multicolumn{1}{l}{FP}    & 8\thinspace bit                                & 6\thinspace bit                        & \multicolumn{1}{l}{FP}             & 8\thinspace bit                         & \multicolumn{1}{l}{FP}    & 8\thinspace bit                       \\
wl$_\text{a}$         & \multicolumn{1}{l}{FP}    & 8\thinspace bit                                & 6\thinspace bit                        & \multicolumn{1}{l}{FP}             & 8\thinspace bit                         & \multicolumn{1}{l}{FP}    & 8\thinspace bit                       \\ \hline
Ours      & \multicolumn{1}{l}{69.69\%} & 69.62                               & \textbf{68.48\%}            & \multicolumn{1}{l}{\textbf{76.00\%}} & \textbf{76.04\%}             & \multicolumn{1}{l}{\textbf{78.26\%}} & \textbf{78.28\%}           \\ \hline
\cite{Nagel2019}    & \textbf{69.7\%}             & \multicolumn{1}{c}{\textbf{69.7\%}} & \multicolumn{1}{c|}{66.3\%} & -                                    & \multicolumn{1}{c|}{-}       & -                           & \multicolumn{1}{c}{-}      \\
\cite{Lee2021}       & -                           & \multicolumn{1}{c}{-}               & \multicolumn{1}{c|}{-}      & -                                    & \multicolumn{1}{c|}{75.68\%} & -                           & \multicolumn{1}{c}{-} \\
\cite{Krishnamoorthi2018}       & -                           & \multicolumn{1}{c}{-}               & \multicolumn{1}{c|}{-}      & 75.2\%                               & \multicolumn{1}{c|}{75.1\%}  & 76.8\%                      & \multicolumn{1}{c}{76.7\%} 
\end{tabular}
\end{table}
\vspace{-0.2cm}

\section{Conclusion}
\label{sec:conclusion}

To fulfill application-specific latency and data security requirements, Deep Neural Networks (DNNs) need to be executed close to the data they are classifying, hence, on edge devices.
Since edge devices offer only a limited memory, the required memory of DNNs has to be reduced.
A common approach for its reduction is limiting the word-length by quantization, which risks decreasing accuracy.
We investigated the impact of word-length of weights and activations on accuracy.
To maintain accuracy, we investigated different quantization options covering varying methods to compute the scaling factor. Quantization is applied post-training to derive various quantized networks on-the-fly from floating-point baseline parameters without the need for computational intensive retraining.
Our investigation indicates the need for higher weight word-lengths compared to activation word-lengths.
Our work highlights the superiority of the scaling factor computation methods AbsMax and AbsP over the state-of-the-art.
With the right combination of quantization options, we achieve higher accuracies with quantized DNNs than with their floating-point counterparts, e.g. for 8\thinspace bit ResNet-50 and ResNet-152.
Finally, we demonstrate that MSE is an inadequate metric to quantify the quality of quantization methods.



\begin{thebibliography}{00}
\bibitem{ImageNetChallenge} O. Russakovsky, J. Deng, H. Su, J. Krause, S. Satheesh, S. Ma, Z. Huang, A. Karpathy, A. Khosla, M. Bernstein, A. C. Berg, F. F. Li, ``ImageNet Large Scale Visual Recognition Challenge,'' In Proceedings of International Journal of Computer Vision (IJCV), 2015, pp. 211--252.
\bibitem{Horowitz2014} M. Horowitz, ``1.1 computing's energy problem (and what we can do about it),'' In IEEE International Solid-State Circuits Conference (IEEE ISSCC), 2014, pp. 10--14.
\bibitem{Latotzke2021} C. Latotzke, and T. Gemmeke, ``Efficiency Versus Accuracy: A Review of design Techniques for DNN Hardware Accelerators,'' In IEEE Access, vol. 9, 2021, pp. 9785--9799.

\bibitem{Malladi2012} K. T. Malladi, F. A. Nothaft, K. Periyathambi, B. C. Lee, C. Kozyrakis, and M. Horowitz, ``Towards energy-proportional datacenter memory with mobile DRAM,’’ In Proceedings of 39th Annual International Symposium on Computer Architecture (ISCA), 2012, pp. 37–-48.
\bibitem{Schaffner2015} M. Schaffner, F. K. Gürkaynak, A. Smolic, and L. Benini, ``DRAM or no-DRAM? Exploring linear solver architectures for image domain warping in 28 nm CMOS,’’ In Proceedings of Design, Automation, \& Test in Europe Conference Exhibition (DATE), May 2015, pp. 707-–712.
\bibitem{Akhlaghi2018} V. Akhlaghi, A. Yazdanbakhsh, K. Samadi, R. K. Gupta, and Hadi Esmaeilzadeh, ``Snapea: Predictive early activation for reducing computation in deep convolutional neural networks,'' In 2018 ACM/IEEE 45th Annual International Symposium on Computer Architecture (ISCA), 2018,  pp. 662--673. 
\bibitem{Clerc2012} S. Clerc, F. Abouzeid, G. Gasiot, D. Gauthier, D. Soussan, and P. Roche, ``A 0.32 V, 55fJ per bit access energy, CMOS 65nm bit-interleaved SRAM with radiation Soft Error tolerance,'' In 2012 IEEE International Conference on IC Design \& Technology, 2012, pp. 1--4.

\bibitem{Mittal2020} S. Mittal, ``A survey of FPGA-based accelerators for convolutional neural networks,'' In Neural Computing and Applications (Neural. Comput. Appl.), vol. 32, no. 4, 2020, pp. 1109--1139.
\bibitem{Yang2019} J. Yang, X. Shen, J. Xing, X. Tian, H. Li, B. Deng, J. Huang, and X. S. Hua, ``Quantization Networks,'' In Proceedings of the IEEE Conference on Computer Vision and Pattern Recognition (CVPR), 2019, pp. 7308--7316.

%
%

\bibitem{Esser2019} S. K. Esser, J. L. McKinstry, D. Bablani, R. Appuswamy, and D. S. Modha, ``Learned step size quantization,'' 2019, arXiv preprint, arXiv:1902.08153.
\bibitem{Bai2021} H. Bai, M. Cao, P. Huang, and J. Shan, ``BatchQuant: Quantized-for-all Architecture Search with Robust Quantizer,'' 2021, arXiv preprint, arXiv:2105.08952.
\bibitem{Bhalgat2020} Y. Bhalgat, J. Lee, M. Nagel, T. Blankevoort, and N. Kwak, ``Lsq+: Improving low-bit quantization through learnable offsets and better initialization,'' In Proceedings of the IEEE/CVF Conference on Computer Vision and Pattern Recognition Workshops, 2020, pp. 696--697.

\bibitem{ResNet} K. He, X. Zhang, S. Ren, and J. Sun, ``Deep residual learning for image recognition,'' In Proceedings of the IEEE conference on computer vision and pattern recognition (CVPR), 2016, pp. 770--778.
\bibitem{ImageNet} J. Deng, W. Dong, R. Socher, J. L. Li, K. Li, and F. F. Li, ``Imagenet: A large-scale hierarchical image database,'' In Proceedings of 2009 IEEE conference on Computer Vision and Pattern Recognition (CVPR), 2009, pp.248--255.

\bibitem{igor} I. Wallossek, ``Nvidia GeForce RTX 2080 Ti im großen Effizienz-Test von 140 bis 340 Watt | igorsLAB,'' https://www.igorslab.de/nvidia-geforce-rtx-2080-ti-im-grossen-effizienz-test-von-140-bis-340-watt-igorslab/, accessed 2022-Sep-06 14:19.
\bibitem{lambdablog} S. Balaban and C. Li, ``RTX 2080 Ti Deep Learning Benchmarks with TensorFlow,'' In https://lambdalabs.com/blog/2080-ti-deep-learning-benchmarks/, accessed 2022-Sep-06 14:19.

\bibitem{Sze2019} Y. H. Chen, T. J. Yang, J. S. Emer, and V. Sze, ``Eyeriss v2: A Flexible Accelerator for Emerging Deep Neural Networks on Mobile Devices,'' In IEEE Journal on Emerging and Selected Topics in Circuits and Systems (IEEE J. Emerg. Sel), vol. 9, no. 2, 2019, pp. 292--308.

\bibitem{Deng2020} L. Deng, G. Li, S. Han, L. Shi, and Y. Xie, ``Model compression and hardware acceleration for neural networks: A comprehensive survey,'' In Proceedings of the IEEE 108, no. 4, 2020, pp. 485--532.
\bibitem{Gholami2021} A. Gholami, S. Kim, Z. Dong, Z. Yao, M. W. Mahoney, and K. Keutzer,``A survey of quantization methods for efficient neural network inference,'' 2021, arXiv preprint arXiv:2103.13630.
\bibitem{Dong2017} X. Dong, S. Chen, and S. Pan, ``Learning to prune deep neural networks via layer-wise optimal brain surgeon,'' In Proceedings of 31st Conference on Neural Information Processing Systems (NIPS), 2017.
\bibitem{Han2016} S. Han, H. Mao, and W. J. Dally, ``Deep compression: Compressing deep neural networks with pruning, trained quantization and huffman coding,'' In Proceedings of International Conference on Learning Representations (ICLR), 2016, http://arxiv.org/abs/1510.00149.

\bibitem{Nogami2019} W. Nogami, T. Ikegami, R. Takano, and T. Kudoh, ``Optimizing Weight Value Quantization for CNN Inference,'' In Proceedings of 2019 International Joint Conference on Neural Networks (IJCNN), 2019, pp. 1--8.
\bibitem{Gupta2020} S. Gupta, S. Ullah, K. Ahuja, A. Tiwari, and A. Kumar, ``ALigN: A Highly Accurate Adaptive Layerwise Log2Lead Quantization of Pre-Trained Neural Networks,'' In IEEE Access, vol. 8, 2020, pp. 118899-118911.

\bibitem{Nagel2020} M. Nagel, R. A. Amjad, M. Van Baalen, C. Louizos, and T. Blankevoort, ``Up or Down? Adaptive Rounding for Post-Training Quantization,'' In Proceedings of International Conference on Machine Learning (ICML), 2020, pp. 7197--7206.

\bibitem{Nagel2019} M. Nagel, M. van Baalen, T. Blankevoort, and M. Welling, ``Data-Free Quantization Through Weight Equalization and Bias Correction,'' In Proceedings of the IEEE/CVF International Conference on Computer Vision (ICCV), 2019, pp. 1325--1334.
\bibitem{Lee2021} D. Lee, M. Cho, S. Lee, J. Song, and C. Choi, ``A Novel Sensitivity Metric For Mixed-Precision Quantization With Synthetic Data Generation,'' In Proceedings of 2021 IEEE International Conference on Image Processing (ICIP), 2021, pp. 1294-1298.

\bibitem{Lin2016} D. Lin, S. Talathi, and S. Annapureddy, ``Fixed point quantization of deep convolutional networks,'' In Proceedings of International Conference on Machine Learning (ICML), 2016, pp. 2849--2858
\bibitem{Zhou2018} Y. Zhou, S. M. Moosavi-Dezfooli, N. M. Cheung, and P. Frossard, ``Adaptive quantization for deep neural network,'' In Proceedings of the 32nd AAAI Conference on Artificial Intelligence (AAAI), 2018, pp. 4596--4604.
\bibitem{Krishnamoorthi2018} R. Krishnamoorthi, ``Quantizing deep convolutional networks for efficient inference: A whitepaper,'' 2018, arXiv preprint arXiv:1806.08342.
\bibitem{Mitschke2019} N. Mitschke, M. Heizmann, K. H. Noffz, and R. Wittmann, ``A Fixed-Point Quantization Technique for Convolutional Neural Networks Based on Weight Scaling,'' In Proceedings of 2019 IEEE International Conference on Image Processing (ICIP), 2019, pp. 3836--3840.
\bibitem{Uhlich2019} S. Uhlich, L. Mauch, F. Cardinaux, K. Yoshiyama, J. A. Garcia, S. Tiedemann, T. Kemp, and A. Nakamura, ``Mixed precision dnns: All you need is a good parametrization,'' 2019, arXiv preprint, arXiv:1905.11452.

\bibitem{Stadtmann2020} T. Stadtmann, C. Latotzke, and T. Gemmeke, ``From Quantitative Analysis to Synthesis of Efficient Binary Neural Networks,'' In Proceedings of the 19th IEEE International Conference On Machine Learning And Applications (ICMLA), 2020, pp. 93--100.
\end{thebibliography}
\end{document}